%% file: 00_main.tex
\documentclass[10pt,twocolumn,letterpaper]{article}

\usepackage{wacv}
\usepackage{times}
\usepackage{epsfig}
\usepackage{graphicx}
\usepackage{multirow}
\usepackage{subcaption}
\usepackage{amsmath}
\usepackage{amssymb}
\usepackage{booktabs}
\def\wacvPaperID{465} % Enter the WACV Paper ID here

\wacvalgorithmstrack   % Uncomment this line if you are submitting to the Algorithms Track.
\wacvfinalcopy % *** Uncomment this line for the final submission

\ifwacvfinal
\usepackage[breaklinks=true,bookmarks=false]{hyperref}
\else
\usepackage[pagebackref=true,breaklinks=true,colorlinks,bookmarks=false]{hyperref}
\fi

\begin{document}

%%%%%%%%% TITLE
\title{CORL: Compositional Representation Learning for Few-Shot Classification}

% \and

\author{Ju He$^1$ \;\; Adam Kortylewski$^{1,2,3}$ \;\;
Alan Yuille$^1$ \\
$^1$Johns Hopkins University \;\; $^2$Max Planck Institute for Informatics \;\;  $^3$University of Freiburg}

\maketitle
\thispagestyle{empty}

\input{01_abstract}
\input{02_introduction}
\input{03_related}

\input{04_method}
\input{05_experiment}
\input{06_conclusion}

{\small
\bibliographystyle{ieee_fullname}
\bibliography{egbib}
}

\end{document}

%% file: 01_abstract.tex
\begin{abstract}
    Few-shot image classification consists of two consecutive learning processes:
    1) In the meta-learning stage, the model acquires a knowledge base from a set of training classes. 2) During meta-testing, the acquired knowledge is used to recognize unseen classes from very few examples.
    Inspired by the compositional representation of objects in humans, we train a neural network architecture that explicitly represents objects as a dictionary of shared components and their spatial composition.
    In particular, during meta-learning, we train a knowledge base that consists of a dictionary of component representations and a dictionary of component activation maps that encode common spatial activation patterns of components. The elements of both dictionaries are shared among the training classes.
    During meta-testing, the representation of unseen classes is learned using the component representations and the component activation maps from the knowledge base.
    Finally, an attention mechanism is used to strengthen those components that are most important for each category. 
    We demonstrate the value of our interpretable compositional learning framework for a few-shot classification using miniImageNet, tieredImageNet, CIFAR-FS, and FC100, where we achieve comparable performance.
\end{abstract}

%% file: 02_introduction.tex
\section{Introduction}

\begin{figure}
    \centering
    \includegraphics[width=\linewidth]{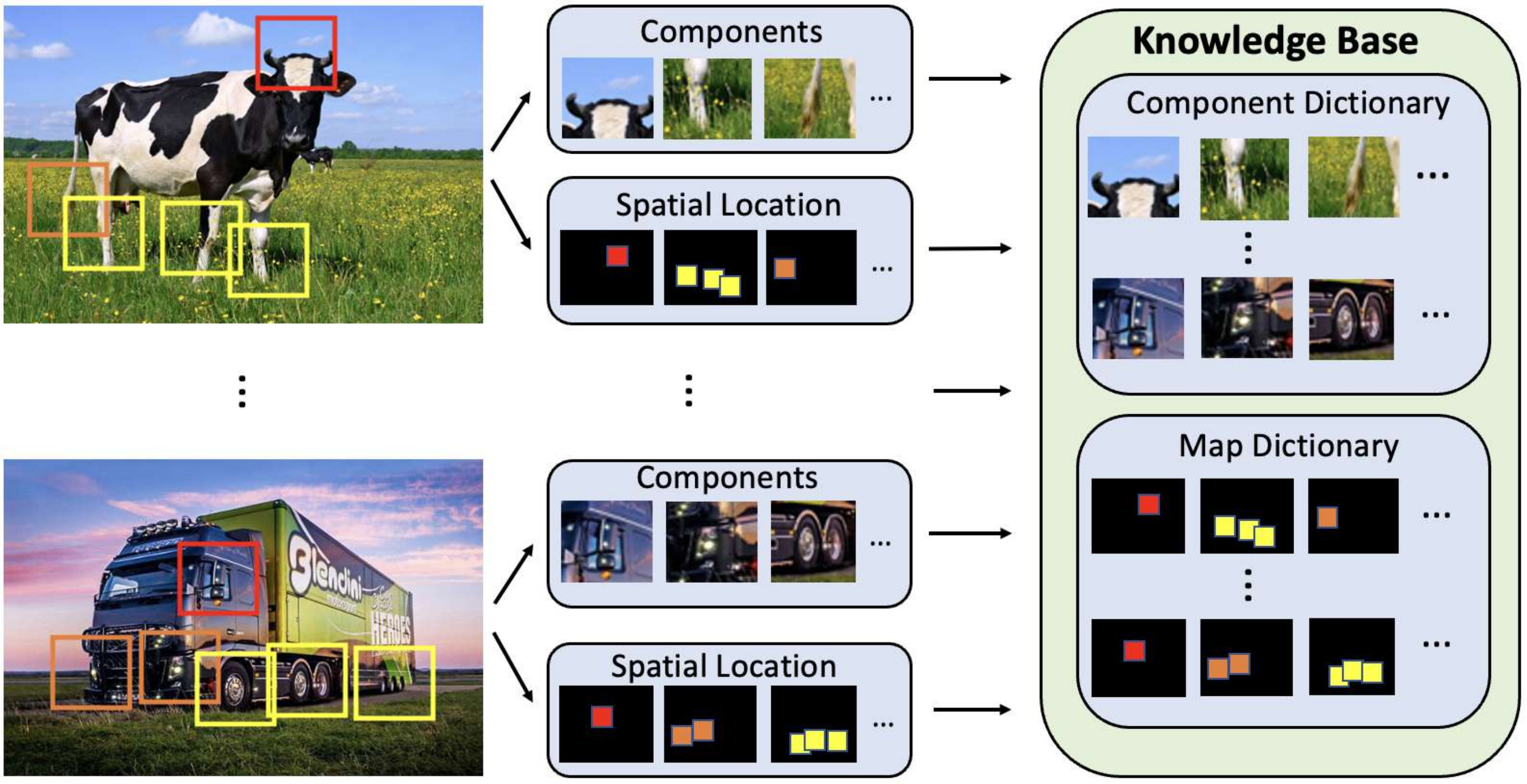}
    \caption{Intuitive illustration of how our model acquires knowledge during meta-learning. In particular, it learns a dictionary of component representations, which resemble individual object components. Some of these can be shared across different classes, e.g., car tires.
    In addition, it learns a map dictionary that contains common spatial activation patterns of components. During meta-testing, the knowledge base facilitates the learning of novel classes by re-using the already learned components and spatial activation patterns.}
    \label{fig:intro}
\end{figure}

Advances in the architecture design of deep convolutional neural networks (DCNNs) \cite{krizhevsky2012imagenet,simonyan2014very,he2016deep} 
increased the performance of computer vision systems at image classification enormously. 
However, in practice, their performance is usually limited when not enough labeled data is available. 
Few-shot classification is concerned with the problem of learning from a small number of samples. 
In particular, it consists of two consecutive learning processes: 1) In the meta-learning stage, the model acquires a knowledge base from a set of training classes. 2) During meta-testing, the acquired knowledge is used to recognize unseen classes from very few examples.
Hence, few-shot classification wants to emulate human learning efficiency \cite{lake2015human, biederman1987recognition, feldman1997structure, jern2013probabilistic} by requiring to transfer the knowledge gained through training on a large number of base classes to enhance the learning of new classes from just a few classes.

Various approaches to few-shot classification were proposed in the past that take different perspectives,
% on the same problem: 1) Meta-learning methods approach the problem from the perspective of learning to learn. They aim to design deep models such that the model parameters can be adapted to include new classes by optimizing very few gradient steps on a small set of new data \cite{finn2017modelagnostic,Nichol2018ReptileAS,Ye_2020_CVPR}. 2) Generalizable embedding methods try to get a more universal feature embeddings that can fit well on both the seen classes and unseen classes which can be easily classified with some simple metrics such as nearest neighbors. \cite{NIPS2017_cb8da676, NIPS2016_90e13578, tian2020rethinking}.
While these methods try to share the common knowledge among base classes and novel classes, since few-shot datasets do not include the attribute or component annotations like in zero-shot datasets, they do not explicitly consider that objects can have similar components and shapes that can be reused. 

In this paper, we introduce a novel approach to few-shot classification that explicitly exploits that object components and their spatial activation patterns can be shared among different object classes. For example, the spatial structure of the class ``horse" can be used for learning efficiently about the class ``donkey".
We implement such a compositional representation sharing by train a knowledge base during meta-learning that consists of a dictionary of component representations and a dictionary of component activation maps that encode common spatial activation patterns of components (Figure \ref{fig:intro}). 
We start by extracting the feature representations of an image up to the last convolution layer of a standard backbone architecture, such as ResNet \cite{he2016deep}.
Following recent work on unsupervised component detection \cite{liao2016learning,Zhang_2018_CVPR,zhang2018deepvoting}, the component dictionary is learned by clustering the individual feature vectors from the feature encoding of the training images.
Moreover, we extract \textit{component activation maps} by computing the spatial activation pattern of components in the training images.
The component activation maps are clustered to learn a dictionary of prototypical maps that encode the most common spatial activation patterns of components.
In practice, the elements of the map dictionary are optimized to be distinct from each other to avoid redundancies.
During meta-testing, our model learns representations of objects by composing them from the components and component activation maps of the knowledge base. 
We use an attention layer to increase the weight of the components that are most discriminative for an object class. 
Finally, the learned object representations are fed into a classifier to predict the class label.
%On top of all these modules, we use fully connected layers to train the whole embedding function when training. 
%And at testing stage, after calculating the attention of different parts, the model already contains enough information and we simply train a SVM classifier based on them using support sets. 
During meta-training, the full model pipeline is trained end-to-end. During meta-testing, we observed that it is sufficient to train the classification head only, while freezing the learned backbone and knowledge base. 
This is different from the majority of other meta-learning methods and highlights the strong generalization performance induced by integrating compositional representation sharing into neural networks.

%In all, our model learns representations by training a neural network as in ordinary classification on the entire meta-training sets which is made by combining all meta-training data into one task. After training, we keep the network's parameters freezed and use the top layers to project the new samples into the embedding space followed by fitting a linear classifier on these embedding features during meta-testing. 
We evaluate our model on four popular few-shot classification datasets and achieve comparable performance on all datasets. 
%Our results verify the effectiveness and superiority of out method when comparing to the state-of-the-art approaches. 
% \begin{comment}
In summary, we make several important contributions in this work:
\begin{enumerate}
    \item  To the best of our knowledge, we are the first to study and demonstrate the effectiveness and interpretability of compositional representation learning on few-shot classification.
    %demonstrate that learning 
    \item We introduce CORL, a novel neural architecture for few-shot classification that implements the inductive prior of compositional representation sharing. It learns a knowledge base with component representations and their common spatial activation patterns, and re-uses this knowledge to learn efficiently about novel classes. 
    \item We achieve comparable performance on several standard benchmarks, outperforming many recent complex optimization methods.
\end{enumerate}
% \end{comment}

%% file: 03_related.tex
\section{Related Work}

In this section, we review existing work on few-shot classification and compositional models.

\subsection{Few-shot learning}

Few-shot learning has received a lot of attention over the last years. Related work can be roughly classified into two branches. The first branch focuses on finding a quick adaptation for the classifier when meeting new tasks. MAML \cite{finn2017model} proposed a general optimization algorithm that can get improvements on a new task with a few gradient steps. MetaOptNet \cite{Lee_2019_CVPR} replaced the linear predictor with an SVM in a MAML framework and introduced a differentiable quadratic programming solver to allow end-to-end training. FEAT \cite{Ye_2020_CVPR} proposed set-to-set functions for a quick adaptation between instance and target embeddings. MELR \cite{fei2020melr} exploited inter-episode relationships to improve the model robustness to poorly-sampled shots. 

Another line of work focuses on how to learn more generalizable feature embeddings and design simple yet efficient metrics for classification. Previous methods mainly learned the representations in a meta-learning framework where the training data was organized in the form of many meta-tasks. Matching Networks \cite{vinyals2016matching} employed two networks for support and query samples, respectively, followed by an LSTM with read-attention to encode the full embedding. Recently, large-training-corpus methods have become the new trend which argue that training a base network on the whole training set directly is also feasible. For example, Dynamic Few-shot \cite{Gidaris_2018_CVPR} extended object recognition systems with an attention weight generator and redesigned the classifier module as the cosine similarity function. RFS \cite{tian2020rethinking} simply trained the embedding function on the combined meta-training sets followed by exploiting knowledge distillation to further improve the performance. It proves that learning a good representation through a proxy task, such as image classification, can give state-of-the-art performances. BML \cite{zhou2021binocular} further improves the learned global feature through an episode-wise local feature learning.

Though all these methods improve few-shot learning in different ways, they do not explicitly take into account that objects can have similar parts and shapes which can be reused. Our method follows the large-training-corpus thought and shows that by explicitly taking into account that objects can have similar components and shapes, we can reuse the information among them and further improve the performance.

\begin{figure*}
    \centering
    \includegraphics[width=\linewidth,height=5cm]{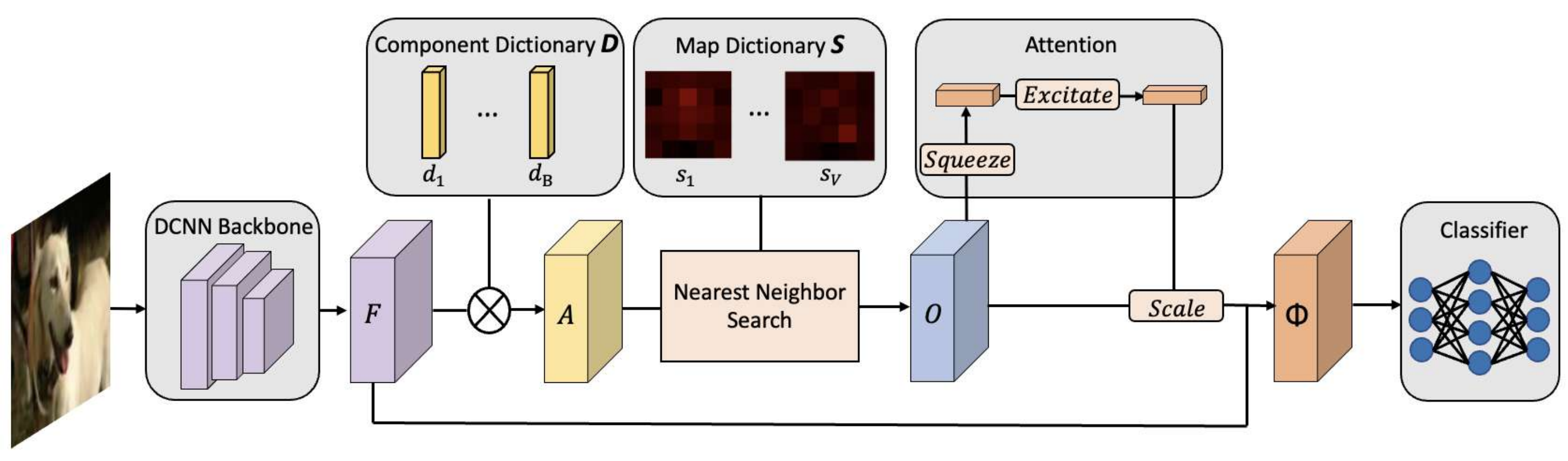}
    \caption{Feed-forward inference with CORL. A DCNN backbone is used to extract the feature map $F$. The items of the component dictionary $D$ are used as kernels to compute a component activation map $A$. We then compare each channel in the component activation map $A_b$ to the spatial patterns in the map dictionary $S$ and multiply it element-wisely with the most similar one to compute the output $O$. An attention mechanism is used to further strengthen components that are most discriminative for an object class. The attention-weighted output is denoted as $\Phi$. We concatenate $\Phi$ with average-pooled $F$ and forward it to the classifier module to compute the final classification result.}
    \label{fig:model}
    \vspace{7pt}
\end{figure*}

\subsection{Compositional models}

%The origin of compositionality in computer vision dated back to Hoffman's research on Parts of Recognition \cite{HOFFMAN198465}. Following this work, models with pictorial structures have been widely studied.

A rich literature on compositional models for image classification exists. However, with the exception of very few works \cite{fidler2007towards, zhu2010part}, most methods use part annotations for training and do not share components among object classes. 
By contrast, our work shows that sharing parts and their spatial distributions among classes without labor-intensive part annotations enables efficient representation learning.
Moreover, many traditional works \cite{fidler2007towards,zhu2010part,Dai_2014_CVPR, fidler2014learning, 1641016} learn the model parameters directly from image pixels.
The major challenge for these approaches is that they need to explicitly account for nuisances such as illumination and deformation. 
Several recent works proposed to learn compositional models from the features of higher layers of deep convolutional neural networks, since these features have shown to be robust to nuisances and have some semantic meanings: 

Liao et al. \cite{liao2016learning} proposed to integrate compositionality into DCNNs by regularizing the feature representations of DCNNs to cluster during learning. Their qualitative results show that the resulting feature clusters resemble detectors of different parts. 
Zhang et al. \cite{Zhang_2018_CVPR} demonstrated that component detectors emerge in DCNNs by restricting the activations in feature maps to have a localized spatial distribution. Kortylewski et al. \cite{kortylewski2019compositional} proposed to learn generative dictionary-based compositional models from the features of a DCNN. They use their compositional model as ``backup" to an independently trained DCNN if the DCNNs classification score falls below a certain threshold. In follow-up work, Kortylewski et al. \cite{Kortylewski_2020_CVPR,kortylewski2020ijcv} further proposed a fully differentiable compositional model for image classification that shows strong robustness to occlusion scenes. 
Sun et al. \cite{sun2020weaklysupervised} demonstrated that these methods could be extended to combine image classification and amodal segmentation by leveraging compositional shape priors.

These recent advances inspire our work in integrating compositional models and deep neural networks.
In this work, we propose to generate part information with compositional model and share it among different classes
In particular, our model for few-shot classification learns component representations and how to compose them together spatially into a whole object representation. 
We exploit that components and their spatial activation patterns can be shared among different classes, which enables our model to learn efficiently from very few examples.
%The parts to  part detectors throughto yield the spatial distribution map and construct a dictionary to store and update these maps. We train the model parameters with back-propagation, while regularizing the compositional model to be generative in terms of the parts.

%% file: 04_method.tex
% \begin{figure*}
%     \centering
%     \includegraphics[width=\linewidth,height=5cm]{pipeline.pdf}
%     \caption{Feed-forward inference with CORL. A DCNN backbone is used to extract the feature map $F$. The items of the component dictionary $D$ are used as kernels to compute a component activation map $A$. We then compare each channel in the component activation map $A_b$ to the spatial patterns in the map dictionary $S$ and multiply it element-wise with the most similar one to compute the output $O$. An attention mechanism is used to further strengthen components that are most discriminative for an object class. The attention-weighted output is denoted as $\Phi$. We concatenate $\Phi$ with average-pooled $F$ and forward it to the classifier module to compute the final classification result.}
%     \label{fig:model}
%     \vspace{7pt}
% \end{figure*}

\section{Method}

We first briefly review the framework of few-shot classification. Then we present how we learn the component dictionary module followed by a discussion on how to learn the map dictionary module and how to integrate these modules into a pipeline for few-shot classification. Lastly, we discuss how to train our model in an end-to-end manner.

\subsection{Few-Shot Classification}
\label{sec:notation}
Few-shot image classification consists of two consecutive learning processes: 1) In the meta-learning stage, the model acquires a knowledge base from a set of training classes. 2) During meta-testing, the acquired knowledge is used to recognize unseen classes from very few examples.
The meta-training set $T$ and meta-testing set $S$ can be both organized as a collection of meta tasks where each meta-task is a N-way-K-shot classification problem.
In this paper, we train our model on the combined meta-training set $T$ followed by directly testing on meta-testing set $S$ without fine-tuning the model parameters.

\subsection{Learning a component dictionary via clustering}
\label{sec:vc}

\noindent \textbf{Formulation.}
We denote a feature map $F^l\in \mathbb{R}^{H\times W \times C}$ as the output of a layer $l$ in a deep convolutional neural network, with $C$ being the number of channels. A feature vector $f^l_p \in \mathbb{R}^C$ is the vector of features in $F^l$ at position $p$ on the 2D lattice $\mathcal{P}$ of the feature map. In the remainder of this section, we omit the superscript $l$ for notational clarity because this is fixed a-priori.

\begin{figure}
    \centering
    \includegraphics[width=\linewidth,height=5cm]{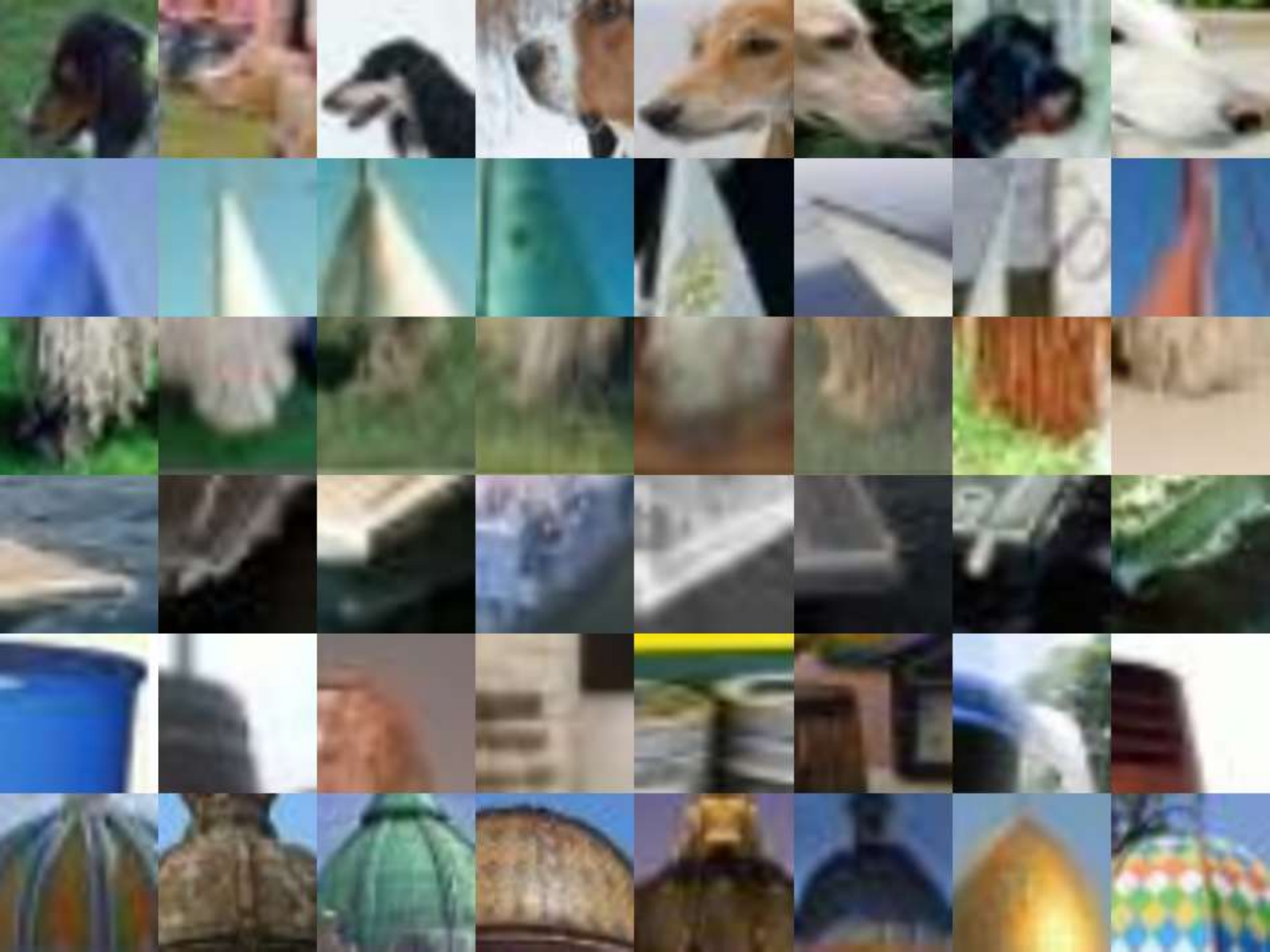}
    \caption{Illustration of the interpretability of our elements in the component dictionary. Each row visualizes image patterns from the miniImageNet dataset that activate a dictionary item the most.}
    \label{fig:vc_example}
\end{figure}

\noindent \textbf{Learning component representations.}
A number of prior works \cite{liao2016learning,Zhang_2018_CVPR,kortylewski2019compositional,Kortylewski_2020_CVPR} on learning compositional representations showed that when clustering feature vectors $f_p$, the cluster centers resemble image patterns that frequently re-occur in the training images. These patterns often share semantic meanings and therefore resemble part-like detectors. Motivated by these results, we aim at constructing a \textit{component dictionary} $D=\{d_1,\dots,d_B\}$, in which the items $d_b \in \mathbb{R}^C$ are cluster centers of the feature vectors $f_p$ from the training images.
To achieve this, we integrate an additional clustering loss which will be introduced later into the overall loss function when training the network. 
Intuitively, this will encourage the dictionary items $d_b$ to learn component representations from the intermediate layer $l$ of a DCNN, and hence to capture the mid-level semantics of objects.
Figure \ref{fig:vc_example} illustrates examples of the dictionary items $d_b$ after the meta-learning stage by showing image patches that activate each item the most. Note how the component representations indeed respond to semantically meaningful image patterns, such as the head of a dog.

\subsection{Compositional Representation Sharing for Few-Shot Classification}
\label{sec:spatial}

\noindent \textbf{Computing the spatial activation maps of components.}
Given the component dictionary $D$, we compute the activation of a component representation $d_b$ at a position $p$ in the feature map $F$ by computing the cosine similarity between $d_b$ and the feature vector $f_p$.
%,\textsl{i.e.},$g(d_h|f^l_P)$. 
We implement this module as a convolution layer, which we call \textit{component detection layer}. %We attach it directly after the normalized intermediate output $F$. 
The convolutional kernels of the component detection layer are the items of the component dictionary $D$, and their kernel size is 1 $\times$ 1. At every forward time, the kernels and input feature maps are $L2$-normalized before computing the cosine similarity.
The output of the detection layer is a component activation tensor $A\in \mathbb{R}^{H \times W \times B}$, where $B$ is the number of items in dictionary $D$. Each channel in this tensor $A_b\in\mathbb{R}^{H\times W}$ is referred to as \textit{component activation map}.
%With this operation, we compute the activation map of each component $d_b$ at each position of the feature map $F$ thus we name $M$ as spatial distribution map. 

\noindent \textbf{Learning dictionaries of spatial activation patterns.}
Our goal is to enable the model to share component activation patterns among different classes. This is inspired by the idea that components of different objects can have similar spatial activation patterns and that this natural redundancy should be exploited (e.g. the spatial structure of the class "dog" can be used for learning efficiently about the class "wolf"). We achieve this by learning a \textit{map dictionary} $\textbf{S}=\{S_1,...,S_V\}$,  which contains the most common component activation patterns in the training data. 
We integrate the dictionary items $S_v\in\mathbb{R}^{H\times W}$ into the feed-forward stage by comparing them to the individual component activation maps $A_b$ using the cosine similarity. We then select the closest item $\bar{v}=\arg\max_v \cos(S_v,A_b)$ and compute the output channel as point-wise multiplication between $A_b$ and $S_{\bar{v}}$. After repeating this operation for all spatial distribution maps, we get the activated spatial distribution output denoted as $O\in\mathbb{R}^{H\times W \times B}$. In this way, each component activation map $A_b$ is encouraged to learn information from the most similar stored spatial activation pattern $S_v$.

\noindent \textbf{Re-weighting important components with attention.}
To further augment components that are most important for representing a particular object, we adopt an attention mechanism to calculate different weights for the spatial distributions. We follow the design of SENet \cite{Hu_2018_CVPR} with small changes. 
In particular, we first squeeze the global spatial information of $O$ into a channel descriptor by using a learned filter $R\in\mathbb{R}^{H\times W \times B}$. Formally, a summary vector $z \in \mathbb{R}^B$ is generated by shrinking $O$ through its spatial dimensions $H \times W$, such that the b-th entry of the vector $z$ is calculated by: 
\begin{align}
    z_b = \sum_{h=1}^H\sum_{w=1}^W R_b(h,w) O_{b}(h,w) \in \mathbb{R}.
\end{align}
%where $R$ refers to squeeze filter.
To fully exploit the squeezed information, we then use the same gating mechanism as 
SENet which contains a bottleneck with two fully-connected layers and non-linearity activation. It can be represented as
\begin{align}
    l = \sigma(W_2\delta(W_1z)) \in\mathbb{R}^B
\end{align}
where $\sigma$ refers to the Sigmoid activation and $W_1$, $W_2$ are the weights of the fully-connected layers. With the computed activation $l$, the final output is obtained by re-weighting the input $O$ with $l$:
\begin{align}
    {\Phi_{b}} = l_{b} \cdot O_{b} \in\mathbb{R}^{H\times W}
\end{align}
where $\cdot$ refers to channel-wise multiplication between the scalar $l_{b}$ and the channel output $O_{b}$.
Finally, we normalize feature vectors along the channel dimension in $\Phi$ to have unit norm and concatenate it with average-pooled $F$ then forward it into the classifier to obtain a final prediction. 

\begin{table*}[ht]
    \small
    \centering
    \setlength{\abovecaptionskip}{0pt}   
    \setlength{\belowcaptionskip}{10pt}
    \caption{\textbf{Comparison to prior work on miniImageNet and tieredImageNet.} Average few-shot classification accuracies(\%) with 95\% confidence intervals on the meta-testing sets of miniImageNet and tieredImageNet. a-b-c-d denotes a 4-layer convolutional network with a, b, c, d filters in each layer.}
    \vspace{10pt} %need to adjust
    \label{tab:imagenet}
    \begin{tabular}{lccccc}
    \hline
    \multicolumn{1}{c}{\multirow{2}{*}{model}} & \multirow{2}{*}{backbone} & \multicolumn{2}{c}{miniImageNet 5-way} & \multicolumn{2}{c}{tieredImageNet 5-way} \\ \cline{3-6} 
    \multicolumn{1}{c}{} &  & 1-shot & 5-shot & 1-shot & 5-shot \\ \hline
    MAML \cite{finn2017model} & 32-32-32-32 & 48.70\ ±\ 1.84 & 63.11\ ±\ 0.92 & 51.67\ ±\ 1.81 & 70.30\ ±\ 1.75 \\
    Matching Networks \cite{vinyals2016matching} & 64-64-64-64 & 43.56\ ±\ 0.84 & 55.31\ ±\ 0.73 & - & - \\
    Prototypical Networks \cite{NIPS2017_cb8da676} & 64-64-64-64 & 49.42\ ±\ 0.78 & 68.20\ ±\ 0.66 & 53.31\ ±\ 0.89 & 72.69\ ±\ 0.74 \\
    Dynamic Few-shot \cite{Gidaris_2018_CVPR} & 64-64-128-128 & 56.20\ ±\ 0.86 & 73.00\ ±\ 0.64 & - & - \\
    Relation Networks \cite{Sung_2018_CVPR} & 64-96-128-256 & 50.44\ ±\ 0.82 & 65.32\ ±\ 0.70 & 54.48\ ±\ 0.93 & 71.32\ ±\ 0.78 \\
    % Shot-Free \cite{Ravichandran_2019_ICCV} & ResNet-12 & 59.04\ ±\ n/a & 77.64\ ±\ n/a & 63.52\ ±\ n/a & 82.59\ ±\ n/a \\
    % TEWAM \cite{Qiao_2019_ICCV} & ResNet-12 & 60.07\ ±\ n/a & 75.90\ ±\ n/a & - & - \\
    % AdaResNet\cite{pmlr-v80-munkhdalai18a} & ResNet-12 & 56.88\ ±\ 0.62 & 71.94\ ±\ 0.57 & - & - \\
    TDADM \cite{NEURIPS2018_66808e32} & ResNet-12 & 58.50\ ±\ 0.30 & 76.70\ ±\ 0.30 & - & - \\
    % Robust 20 \cite{dvornik2019diversity} & ResNet-12 & 59.38\ ±\ 0.65 & 76.90\ ±\ 0.42 & - & - \\
    % Variational FSL \cite{Zhang_2019_ICCV} & ResNet-12 & 61.23\ ±\ 0.26 & 77.69\ ±\ 0.17 & - & - \\
    % LEO-trainval \cite{rusu2018meta} & WRN-28-10 & 61.76\ ±\ 0.08 & 77.59\ ±\ 0.12 & 66.33\ ±\ 0.05 & 81.44\ ±\ 0.09 \\
    MetaOptNet \cite{Lee_2019_CVPR} & ResNet-12 & 62.64\ ±\ 0.61 & 78.63\ ±\ 0.46 & 65.99\ ±\ 0.72 & 81.56\ ±\ 0.53 \\
    FEAT \cite{Ye_2020_CVPR} & WRN-28-10 & 65.10\ ±\ 0.20 & 81.11\ ±\ 0.14 & 70.41\ ±\ 0.23 & 84.38\ ±\ 0.16 \\
    RFS \cite{tian2020rethinking} & ResNet-12 & 64.82\ ±\ 0.60 & 82.14\ ±\ 0.43 & 71.52\ ±\ 0.69 & 86.03\ ±\ 0.49 \\
    Neg-Cosine \cite{liu2020negative}& ResNet-12 & 63.85\ ±\ 0.81 & 81.57\ ±\ 0.56 & - & - \\
    MABAS \cite{kim2020model} & ResNet-12 & 65.08\ ±\ 0.86 & 82.70\ ±\ 0.54 & \textbf{74.40\ ±\ 0.68} & 86.61\ ±\ 0.59 \\
    MELR \cite{fei2020melr} & ResNet-12 & 67.40\ ±\ 0.43 & 83.40\ ±\ 0.28 & 72.14\ ±\ 0.51 & 87.01\ ±\ 0.35 \\
    BML \cite{zhou2021binocular} & ResNet-12 & 67.04\ ±\ 0.63 & 83.63\ ±\ 0.29 & 68.99\ ±\ 0.50 & 85.49\ ±\ 0.34 \\
    DMF \cite{xu2021learning} & ResNet-12 & \textbf{67.76\ ±\ 0.46} & 82.71\ ±\ 0.31 & 71.89\ ±\ 0.52 & 85.96\ ±\ 0.35 \\
    infoPatch \cite{liu2021learning} & ResNet-12 & 67.67\ ±\ 0.45 & 82.44\ ±\ 0.31 & 71.51\ ±\ 0.52 & 85.44\ ±\ 0.35 \\
    IEPT \cite{zhang2020iept} & ResNet-12 & 67.05\ ±\ 0.44 & 82.90\ ±\ 0.30 & 72.24\ ±\ 0.50 & 86.73\ ±\ 0.34 \\
    Meta DeepBDC \cite{xie2022joint} & ResNet-12 & 67.34\ ±\ 0.43 & \textbf{84.46\ ±\ 0.28} & 72.34\ ±\ 0.49 & \textbf{87.31\ ±\ 0.32} \\ \hline
    Ours & ResNet-12 & 65.74\ ±\ 0.53 & 83.03\ ±\ 0.33 & 73.82\ ±\ 0.58 & 86.76\ ±\ 0.52\\ \hline
    \end{tabular}
\end{table*}

\subsection{End-to-end Training of the model}
\label{sec:train}

During training, we use a two-layer fully-connected structure as a classifier to predict the classification results. Our model is fully differentiable and can be trained end-to-end using back-propagation. The trainable parameters of our model are $\Theta = \{\Omega,D, S\}$, where $\Omega$ are the parameters of the backbone used for feature extraction, e.g., ResNet-12. $D$ is the component dictionary, and $S$ is the dictionary of component activation maps. 
We optimize these parameters jointly using stochastic gradient descent. Our loss function contains three terms:
\begin{equation}
\label{equ:loss}
    \begin{aligned}
        \mathcal{L}(y, y') = \mathcal{L}_{class}(y, y') + \gamma_{1}\mathcal{L}_{cluster}(D) + \\
        \gamma_{2}\mathcal{L}_{sparse}(S) 
\end{aligned}
\end{equation}

$\mathcal{L}_{class}(y,y')$ is the cross-entropy loss between the predicted label $y'$ and the ground-truth label $y$.
The second term $\mathcal{L}_{cluster}(D)$ is used to add additional regularization for the dictionary of components:
\begin{align}
    \mathcal{L}_{cluster}(D) = \sum\limits_{p}\min\limits_{b}(1-\cos(D_b|f_p))
\end{align}
where $f_p$ refers to the feature vector at position $p$ in the feature map $F$ and $\cos(\cdot,\cdot)$ refers to the cosine similarity. Intuitively, this loss encourages the dictionary's items to become similar to the feature vectors $f_p$. Thus the dictionary is forced to learn component representations that frequently occur in the training data.

To regularize the map dictionary, we add a sparse loss on the dictionary $S$:
\begin{align}
    \mathcal{L}_{sparse} = \sum\limits_{v=1}^{V}\arg\max\limits_{v'}\cos(S_v,S_{v'})^2
\end{align}
where $\cos(S_v,S_{v'})$ is the cosine similarity between two dictionary elements of $S$.
This regularizer encourages the map dictionary elements to be sparse, thus avoiding that the elements become too similar to each other. We find that exploiting the second-order information of the cosine similarity avoids that the sparse loss will dominate the direction of the gradient at later stages of the training and thus helps the model to converge.

\subsection{Replacing the classifier during meta-testing}
At meta-testing time, unlike many other methods, we do not further fine-tune our model based on the support sets $D_{j}^{train}$ in the meta-testing stage. Instead, we replace the fully-connected classification head with a simpler classifier 
%some distance measurement or a linear classifier head since training a new fully-connected layer on the support set will lead 
to avoid overfitting. 
We tested different classifier, such as nearest neighbor based on different distance metrics, logistic regression classifier, linear support vector machine. We found that the logistic regression gives the best results. In summary, for a task $(D_{j}^{train}, D_{j}^{test})$ sampled from meta-testing set $S$, we forward $D_{j}^{train}$ through the whole embedding function to get the attentioned component activation map $\Phi$ contacted with the average-pooled $F$, and train the logistic regression classifier on this representation.

%% file: 05_experiment.tex
\section{Experiment}

In this section, we conduct extensive experiments that prove the effectiveness of our model. 
We first describe our detailed setup, which includes datasets, model structure, and hyper-parameters. 
Then we evaluate our model and make comparisons to related work on four few-shot classification benchmark datasets: miniImageNet \cite{vinyals2016matching}, tieredImageNet \cite{ren2018meta}, CIFAR-FS \cite{bertinetto2018meta}, Fewshot-CIFAR100 (FC100) \cite{NEURIPS2018_66808e32}. 
The concrete performance on ImageNet and CIFAR derivatives are discussed respectively. 
We further conduct ablation studies to study the effects of the individual modules in our CORL pipeline.
In the end, we show the receptive field of items in our component dictionary.

\subsection{Experimental Setups}
\label{sec:setup}

\textbf{Architecture.} Following previous work \cite{mishra2017simple, NEURIPS2018_66808e32, Ravichandran_2019_ICCV, dhillon2019baseline}, we use a ResNet12 as our feature extraction network which contains 4 residual blocks, where each of them contains 3 convolution layers. We drop the last average-pooling layer and use feature maps before the pooling for later computation. 
Dropblock is used in our model as a regularizer. The number of items in the component dictionary $D$ is 512 and the number of items in the map dictionary $S$ is 2048.

\noindent \textbf{Implementation details.} The loss coefficients in Eq. \ref{equ:loss} are set to $\gamma_1=1$ and $\gamma_2=0.5$ respectively. We use the SGD optimizer with a momentum of $0.9$ and a weight decay of $5e^{-4}$. Our batch size is set to $64$, and the base learning rate is $0.05$. 
We initialize the component dictionary $D$ via K-means clustering on the feature vectors $f_p$ and fine-tune it at the meta-training stage.
We found that a random initialization of the component dictionary would not reduce the final performance, but the K-means initialization helps our model to converge faster as the cluster loss is lower at the start of training. 
On miniImageNet and tieredImageNet, we train our model $100$ epochs and for CIFAR derivatives, the total epochs for training are $90$. We adopt cosine annealing as the learning rate scheduler. During training, we adopt regular data augmentation schemes such as random flipping. When handling CIFAR derivatives datasets, we resize the input image to $84 \times 84$ pixels in order to have enough spatial resolution. 
Following common experimental setups, we report our performance based on an average of $600$ meta-tasks, where each of them contains 15 test instances per class.
For fair comparison, we only train our model on the training set of each dataset and do not perform any test-time training.

\subsection{Experiments on ImageNet derivatives}
\label{sec:IN}

\noindent \textbf{The miniImageNet dataset} is the most classic few-shot classification benchmark proposed by Matching Networks \cite{vinyals2016matching}. It consists of 100 randomly sampled different classes, and each class contains 600 images of size 84 $\times$ 84 pixels. We follow the widely-used splitting protocol proposed by Ravi et al. \cite{ravi2016optimization}, which uses 64 classes for meta-training, 16 classes for meta-validation, and 20 classes for meta-testing.

\noindent \textbf{The tieredImageNet dataset} is a larger subset of ImageNet, composed of 608 classes grouped into 34 high-level categories. They are further divided into 20 categories for training, 6 categories for validation, and 8 categories for testing, which corresponds to 351, 97, and 160 classes for meta-training, meta-validation, and meta-testing, respectively. This splitting method, which considers high-level categories, is applied to minimize the semantic overlap between the splits. Images are of size 84 $\times$ 84.

\begin{table*}[ht]
    \small
    \centering
    \setlength{\abovecaptionskip}{0pt}   
    \setlength{\belowcaptionskip}{10pt}
    \caption{\textbf{Comparison to prior work on CIFAR-FS and FC100.} Average few-shot classification accuracies(\%) with 95\% confidence intervals on the meta-testing sets of CIFAR-FS and FC100. a-b-c-d denotes a 4-layer convolutional network with a, b, c, d filters in each layer.}
    \vspace{10pt} %need to adjust
    \label{tab:cifar}
    \begin{tabular}{lccccc}
    \hline
    \multicolumn{1}{c}{\multirow{2}{*}{model}} & \multirow{2}{*}{backbone} & \multicolumn{2}{c}{CIFAR-FS 5-way} & \multicolumn{2}{c}{FC100 5-way} \\ \cline{3-6} 
    \multicolumn{1}{c}{} &  & 1-shot & 5-shot & 1-shot & 5-shot \\ \hline
    MAML \cite{finn2017model} & 32-32-32-32 & 58.90\ ±\ 1.90 & 71.50\ ±\ 1.00 & - & - \\
    Prototypical Networks \cite{NIPS2017_cb8da676} & 64-64-64-64 & 55.50\ ±\ 0.70 & 72.00\ ±\ 0.60 & 35.30\ ±\ 0.60 & 48.60\ ±\ 0.60 \\
    Relation Networks \cite{Sung_2018_CVPR} &  64-96-128-256 & 55.00\ ±\ 1.00 & 69.30\ ±\ 0.80 & - & - \\
    % R2D2 \cite{bertinetto2018meta} & 96-192-384-512 & 65.30\ ±\ 0.20 & 79.40\ ±\ 0.10 & - & - \\
    TADAM \cite{NEURIPS2018_66808e32} & ResNet-12 & - & - & 40.10\ ±\ 0.40 & 56.10\ ±\ 0.40 \\
    Shot-Free \cite{Ravichandran_2019_ICCV} & ResNet-12 & 69.20\ ±\ n/a & 84.70\ ±\ n/a & - & - \\
    TEWAM \cite{Qiao_2019_ICCV} & ResNet-12 & 70.40\ ±\ n/a & 81.30\ ±\ n/a & - & - \\
    Prototypical Networks \cite{NIPS2017_cb8da676} & ResNet-12 & 72.20\ ±\ 0.70 & 83.50\ ±\ 0.50 & 37.50\ ±\ 0.60 & 52.50\ ±\ 0.60 \\
    MetaOptNet \cite{Lee_2019_CVPR} & ResNet-12 & 72.60\ ±\ 0.70 & 84.30\ ±\ 0.50 & 41.10\ ±\ 0.60 & 55.50\ ±\ 0.60 \\
    DeepEMD \cite{Zhang_2020_CVPR} & ResNet-12 & - & - & \textbf{46.47\ ±\ 0.78} & \textbf{63.22\ ±\ 0.71} \\
    RFS \cite{tian2020rethinking} & ResNet-12 & 73.90\ ±\ 0.80 & 86.90\ ±\ 0.50 & 44.60\ ±\ 0.70 & 60.90\ ±\ 0.60 \\
    MABAS \cite{kim2020model} & ResNet-12 & 73.51\ ±\ 0.96 & 85.49\ ±\ 0.68 & 42.31\ ±\ 0.75 & 57.56\ ±\ 0.78 \\
    ConstellationNet \cite{xu2020attentional} & ResNet-12 & \textbf{75.40\ ±\ 0.20} & 86.80\ ±\ 0.20 & 43.80\ ±\ 0.20 & 59.70\ ±\ 0.20 \\
    BML \cite{zhou2021binocular} & ResNet-12 & 73.45\ ±\ 0.47 & 88.04\ ±\ 0.33 & - & - \\ \hline
    Ours & ResNet-12 & 74.13\ ±\ 0.71 & \textbf{87.54\ ±\ 0.51} & 44.82\ ±\ 0.73 & 61.31\ ±\ 0.54 \\ \hline
    \end{tabular}
\end{table*}

\begin{table*}[]
    \small
    \centering
    \setlength{\abovecaptionskip}{0pt}   
    \setlength{\belowcaptionskip}{10pt}
    \caption{\textbf{Ablation study.} Performance of our ablated models on four few-shot classification benchmarks. The model is fixed when we conduct experiments about ablation for loss terms. The metric is average few-shot classification accuracies(\%).}
    \vspace{10pt} %need to adjust
    \label{tab:ablation}
    \begin{tabular}{cccc|cccccccc}
    \hline
    \multirow{2}{*}{Map Dictionary} & \multirow{2}{*}{Attention} & \multirow{2}{*}{Cluster Loss} & \multirow{2}{*}{Sparse Loss} & \multicolumn{2}{c}{miniImageNet} & \multicolumn{2}{c}{tieredImageNet} & \multicolumn{2}{c}{CIFAR-FS} & \multicolumn{2}{c}{FC100} \\ \cline{5-12} 
    &  &  &  & 1-shot & 5-shot & 1-shot & 5-shot & 1-shot & 5-shot & 1-shot & 5-shot \\ \hline
     &  &  &  & 61.42 & 77.43 & 69.67 & 82.33 & 70.23 & 83.21 & 40.28 & 57.02 \\
    \checkmark &  &  &  & 61.71 & 78.65 & 70.53 & 82.67 & 70.89 & 83.72 & 40.62 & 57.72 \\
    \checkmark & \checkmark &  &  & 62.03 & 79.44 & 71.10 & 83.42 & 71.67 & 84.59 & 41.13 & 58.25 \\
    \checkmark & \checkmark & \checkmark & & 64.24 & 81.74 & 72.32 & 85.93 & 73.07 & 85.54 & 43.24 & 60.35 \\
    \checkmark & \checkmark & \checkmark & \checkmark & \textbf{65.74} & \textbf{83.03} &  \textbf{73.82} &  \textbf{86.76} & \textbf{74.13} & \textbf{87.54} & \textbf{44.82} & \textbf{61.31} \\ \hline
    \end{tabular}
\end{table*}

\begin{figure*}[h]
    \centering
    \begin{subfigure}[]{0.49\linewidth}
         \includegraphics[width=\linewidth]{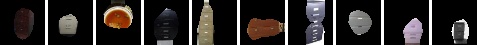}
     \end{subfigure}
    \begin{subfigure}[]{0.49\linewidth}
         \includegraphics[width=\linewidth]{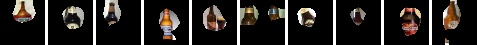}
     \end{subfigure}
     \begin{subfigure}[]{0.49\linewidth}
         \includegraphics[width=\linewidth]{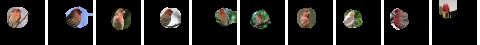}
     \end{subfigure}
    \begin{subfigure}[]{0.49\linewidth}
         \includegraphics[width=\linewidth]{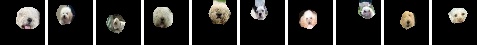}
     \end{subfigure}
     \begin{subfigure}[]{0.49\linewidth}
         \includegraphics[width=\linewidth]{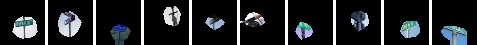}
     \end{subfigure}
    \begin{subfigure}[]{0.49\linewidth}
         \includegraphics[width=\linewidth]{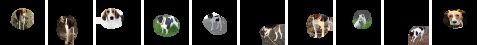}
     \end{subfigure}
     \begin{subfigure}[]{0.49\linewidth}
         \includegraphics[width=\linewidth]{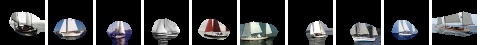}
     \end{subfigure}
    \begin{subfigure}[]{0.49\linewidth}
         \includegraphics[width=\linewidth]{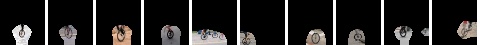}
     \end{subfigure}
     \begin{subfigure}[]{0.49\linewidth}
         \includegraphics[width=\linewidth]{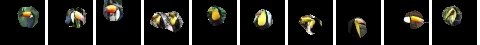}
     \end{subfigure}
    \begin{subfigure}[]{0.49\linewidth}
         \includegraphics[width=\linewidth]{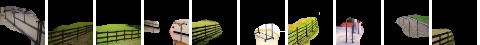}
     \end{subfigure}
     \begin{subfigure}[]{0.49\linewidth}
         \includegraphics[width=\linewidth]{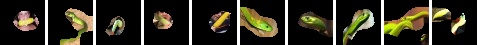}
     \end{subfigure}
    \begin{subfigure}[]{0.49\linewidth}
         \includegraphics[width=\linewidth]{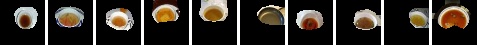}
     \end{subfigure}
     \caption{\textbf{Visualization of elements in the component dictionary $D$.} Each row visualizes the activation of two dictionary components. Note that the learned components activate to semantically meaningful image patterns such as tires, animal heads, or parts of a bottle, even though no part annotations are used during training process.}
     \label{fig:vc_vis}
\end{figure*}

\noindent \textbf{Results.} Table \ref{tab:imagenet} summarizes the results on the 5-way miniImageNet and tieredImageNet. Our method achieves comparable performance on the miniImageNet benchmark for both 5-way-1-shot and 5-way-5-shot tasks. On tieredImageNet, we also achieve the best performance on the 5-way-5-shot task and comparable performance on the 5-way-1-shot task. 
Note that related works use very complex training schemes to improve their performance. For example, LEO \cite{rusu2018meta} used an encoder and relation network in addition to the WRN-28-10 backbone network to produce sample-depend initialization of the gradient descent. FEAT \cite{Ye_2020_CVPR} and LEO \cite{rusu2018meta} pre-train the WRN-28-10 backbone to classify 64 meta-training set of miniImageNet and then continue meta-training.
% TADAM \cite{NEURIPS2018_66808e32} co-trained the feature embedding on both meta-training task (5-way) and the standard classification task (64-way) together. 
FEAT \cite{Ye_2020_CVPR} and MABAS \cite{kim2020model} require additional fine-tuning on meta-testing sets. 
In contrast to all those approaches, our model just needs to train the embedding function through standard classification without further fine-tuning. This strategy allows us to clearly demonstrate the effect of a good embedding function by achieving stronger performance with an arguably simpler training. 
% We also notice that when comparing to the best performance on each benchmark, the performance gain for miniImageNet is relatively larger than that of tieredImageNet.

\subsection{Experiments on CIFAR derivatives}
\label{sec:CF}

\noindent \textbf{The CIFAR-FS dataset} is a recently proposed few-shot image classification benchmark derived from CIFAR. It consists of all 100 classes and is further randomly split into 64 training classes, 16 validation classes, and 20 testing classes. Each class contains 600 images of size 32 $\times$ 32.

\noindent \textbf{The FC100 dataset} is another few-shot classification dataset based on CIFAR. Its main idea is very similar to tieredImageNet, where the whole 100 classes are grouped into 20 superclasses. Each superclass is composed of standard 5 classes. These superclasses are divided into 12, 4, 4 for training, validation, testing correspondingly. 
% Images are of size 32 $\times$ 32 pixels.

\noindent \textbf{Results.} Table \ref{tab:cifar} summarizes the performance on the 5-way CIFAR-FS and FC100. Our model achieves comparable performance on all tasks in both CIFAR-FS and FC100 benchmark.
We observe that the relative improvement rate on the CIFAR-FS dataset is larger compared to the FC100 dataset which is similar to generalization pattern on the ImageNet derivatives. Namely, the performance on the benchmark with semantic gaps between the meta-training set and meta-testing set benefits less from our method. We expect to alleviate this problem by finding a good way to fine-tune our model at meta-testing stage in future work.

\subsection{Ablation Experiments}
\label{sec:ablation}

In this section, we conduct ablation studies on our CORL pipeline to analyze how its variants affect the few-shot classification result. 
We study the following three parts of our method: (a) The map dictionary; (b) The attention module on activated spatial distribution maps; (c) The cluster loss of the component dictionary; (d) The sparse loss of the map dictionary. 
In addition, we also analyze the result of the number of items in the component dictionary $D$, map dictionary $S$.

Table \ref{tab:ablation} shows the result of our ablation studies on miniImageNet, tieredImageNet, CIFAR-FS and FC100. We can see that when introducing the map dictionary, the model goes beyond the pure bag-of-words model and achieves on average 0.7\% performance gain even without further using loss to restrain it. This clearly shows that considering the spatial relationship between components helps the model. Besides, the attention mechanism for augmenting important components and their relationship makes the average performance improve on average around $0.6\%$ on all datasets. With our cluster loss that regularizes the items in the component dictionary $D$, we gain on average about $2.3\%$. In addition, this loss increases the interpretability of our model as it makes the image patches detected by these component detectors more semantically meaningful. 
Our sparse loss regularizer improves the performance by another $1.5\%$, which demonstrates the benefit of making the items in the map dictionary distinct from each other.% help to accurately activate the specific input spatial distribution map. 

\begin{table}
    \small
    \centering
    \setlength{\abovecaptionskip}{0pt}   
    \setlength{\belowcaptionskip}{10pt}
    \caption{\textbf{Test accuracies(\%) on meta-testing set of miniImageNet with a varying number of items in the component dictionary.} Either too much or too less items harm the performance of the model.}
    \vspace{10pt} %need to adjust
    \label{tab:part}
    \begin{tabular}{c|cc}
    \hline
    \multirow{2}{*}{Component Dictionary Size B} & \multicolumn{2}{c}{miniImageNet} \\ \cline{2-3} 
     & 1-shot & 5-shot \\ \hline
    256 & 63.82 & 81.13 \\
    512 & \textbf{65.74} & \textbf{83.03} \\
    1024 & 65.12 & 82.45 \\ \hline
    \end{tabular}
\end{table}

Table \ref{tab:part} shows the influence of the size $B$ of the component dictionary $D$ on the performance of our model on miniImageNet. With too less items in the dictionary, our model do not contain enough information for modeling the component-whole based relationships of the objects. However, if the size $B$ becomes too large, it harms each component representation to accurately capture the corresponding features and many items might focus on meaningless background thus enlarge the learning difficulty.

\begin{figure}
    \centering
    \includegraphics[width=\linewidth,height=3cm]{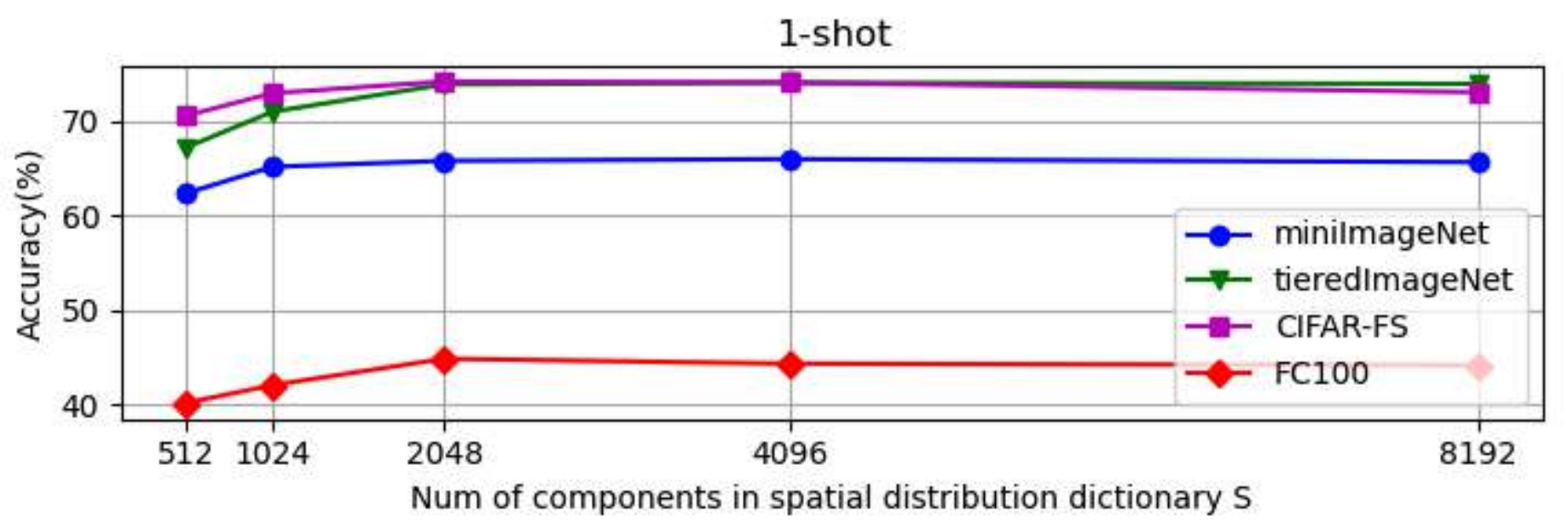}
    \caption{\textbf{Test accuracies(\%) on meta-testing sets with a varying number of items in the map dictionary.} The performance of our model increases at first and saturates at some point with a slight tendency to drop further.}
    \label{fig:num}
\end{figure}

Figure \ref{fig:num} illustrates the influence of the number of items in the map dictionary $S$ on the performance of our model on four benchmarks. 
The performance improves at first when the number of items increases but saturates as the dictionaries become larger. The performance keeps at the same level and even shows a tendency to drop. 
These results suggest that when the capacity of the dictionary is small, our model cannot store all necessary information. 
However, if the capacity becomes too large, the model starts overfitting.% each component can only learn a rough representation without getting enough updates.
% \begin{figure}
%     \centering
%     \includegraphics[width=\linewidth,height=5cm]{tsne.pdf}
%     \caption{\textbf{t-SNE visualization illustrating improved feature embeddings with our designed modules.} The left figure corresponds to our model without attention module, cluster loss, and sparse loss. The right figure corresponds to the complete model.}
%     \label{fig:tsne}
% \end{figure}

% In Figure \ref{fig:tsne}, we further show the t-SNE visualization results of the embedding space of our model on the CIFAR-FS dataset. For both t-SNE plots, we use the same data and the same hyper-parameters. 
% The left figure shows the embedding space without using attention, the cluster, and sparse loss for regularization. The right figure shows the result with the full model. 
% We can observe that even without these modules, the initial embedding is quite good which can be attributed to the novel architecture design. However, after adding these modules, the cluster centers are forced to become even denser and more distinct from each other.

\subsection{Visualization of the components}
\label{sec:comvis}
In Figure \ref{fig:vc_vis} we visualize the activation of the elements in the component dictionary $D$ following the method proposed by \cite{zhou2014object}. Specifically, we threshold the response map of the dictionary elements on images from the miniImageNet dataset and scale them up to the original image size. The visualization results show that the dictionary components respond to semantically meaningful image patterns such as tires, animal heads, or parts of a bottle. 
In summary, the strong discriminative abilities of our model (Tables \ref{tab:imagenet} \& \ref{tab:cifar}) and the qualitative visualization in Figure \ref{fig:vc_vis} suggests that our compositional representation learning approach enables the model to learn semantically meaningful local image patterns with no part-level supervisions during training.

%% file: 06_conclusion.tex
\section{Conclusion}
In this work, we study the problem of few-shot image classification. 
Inspired by the compositional representation of objects in humans, we introduce CORL, a novel neural architecture for few-shot classification that learns through compositional representation sharing.
In particular, CORL learns a knowledge base that contains a dictionary of component representations and a dictionary of component activation maps that encode frequent spatial activation patterns of components. During meta-testing, this knowledge is reused to learn about unseen classes from very few samples.
Our extensive experiments demonstrate the effectiveness of our method, which achieves comparable performance on four popular few-shot classification benchmarks.

\textbf{Acknowledgements.} The authors gratefully acknowledge supports from ONR N00014-21-1-2690. AK acknowledges support via his Emmy Noether Research Group funded by the German Science Foundation (DFG) under Grant No. 468670075.